# Value Function Approximation in Noisy Environments Using Locally Smoothed Regularized Approximate Linear Programs

**Gavin Taylor**
Department of Computer Science
United States Naval Academy
Annapolis, MD 21402

**Ronald Parr**
Department of Computer Science
Duke University
Durham, NC 27708

## Abstract

Recently, Petrik et al. demonstrated that $L_1$-Regularized Approximate Linear Programming (RALP) could produce value functions and policies which compared favorably to established linear value function approximation techniques like LSPI. RALP's success primarily stems from the ability to solve the feature selection and value function approximation steps simultaneously. RALP's performance guarantees become looser if sampled next states are used. For very noisy domains, RALP requires an accurate model rather than samples, which can be unrealistic in some practical scenarios. In this paper, we demonstrate this weakness, and then introduce Locally Smoothed $L_1$-Regularized Approximate Linear Programming (LS-RALP). We demonstrate that LS-RALP mitigates inaccuracies stemming from noise even without an accurate model. We show that, given some smoothness assumptions, as the number of samples increases, error from noise approaches zero, and provide experimental examples of LS-RALP's success on common reinforcement learning benchmark problems.

## 1 Introduction

Markov Decision Processes (MDPs) are applicable to a large number of real-world Artificial Intelligence problems. Unfortunately, solving large MDPs is computationally challenging, and it is generally accepted that such MDPs can only be solved approximately. This approximation is often done by relying on linear value function approximation, in which the value function is chosen from a low-dimensional vector space of features. Generally speaking, agent interactions with the environment are sampled, the vector space is defined through feature selection, a value function is selected from this space by weighting features, and finally a policy is extracted from the value function. Each of these steps contains difficult problems and is an area of active research. The long-term hope for the community is to handle each of these topics on increasingly difficult domains with decreasing requirements for expert guidance (providing a model, features, etc.) in application. This will open the door to wider application of reinforcement learning techniques.

Previously, Petrik et al. [12] introduced $L_1$-Regularized Approximate Linear Programming (RALP) to combine and automate the feature selection and feature weighting steps of this process. Because RALP addresses these steps in tandem, it is able to take advantage of extremely rich feature spaces without overfitting. However, RALP's hard constraints leave it especially vulnerable to errors when noisy, sampled next states are used. This is because each sampled next state generates a lower bound on the value of the originating state. Without a model, the LP will not use the expectation of these next state values but will instead use the max of the union of the discounted values of sampled next states. Thus, a single low probability transition can determine the value assigned to a state when samples are used in lieu of a model. This inability to handle noisy samples is a key limitation of LP based approaches.

In this paper, we present Locally Smoothed $L_1$-Regularized Approximate Linear Programming (LS-RALP), in which a smoothing function is applied to the constraints of RALP. We show this smoothing function reduces the effects of noise, without the requirement of a model. The result is a technique which accepts samples and a large candidate feature set, and then automatically chooses features and approximates the value function, requiring neither a model, nor human-guided feature engineering.

Standard ALP approaches produce value functions and require a model at policy execution time to determine the optimal action with respect the value function returned by the LP. We extend LS-RALP to produce Q-functions, rather than value functions, thereby removing the dependence on a model from both value function learning and policy execution.

After introducing notation, this paper first examines RALP in Section 3 to demonstrate the effect of noise on solution quality. Section 4 then introduces LS-RALP, and explains the approach. In Section 5, we prove that as samples are added, the error from noise in LS-RALP's value function approximation approaches zero. Finally, in Section 6, we empirically demonstrate LS-RALP's success in high-noise environments, even with a small number of samples. Finally, in Section 7, we make our concluding remarks.

## 2 Framework and Notation

In this section, we formally define Markov decision processes and linear value function approximation. A *Markov decision process* (MDP) is a tuple $(\mathcal{S}, \mathcal{A}, p, R, \gamma)$, where $\mathcal{S}$ is a measurable subset of Euclidean space, and $\mathcal{A}$ is a finite set of actions. $p$ is a transition kernel, where $p(s'|s, a)$ represents the conditional density for next states. The function $R : \mathcal{S} \mapsto \Re$ is a measurable reward function, and $\gamma$ is a discount factor.

We are concerned with finding a value function $V$ that maps each state $s \in \mathcal{S}$ to the expected total $\gamma$-discounted reward for the process. Value functions can be useful in creating or analyzing a measurable policy function $\pi : \mathcal{S} \times \mathcal{A} \to [0,1]$ such that for all $s \in \mathcal{S}, \sum_{a \in \mathcal{A}} \pi(s, a) = 1$. The transition and reward functions for a given policy are denoted by $P_\pi$ and $R_\pi$. We denote the Bellman operator for a given policy as $\bar{T}_\pi$, and the max Bellman operator simply as $\bar{T}$. That is, for some state $s \in \mathcal{S}$:

$$\bar{T}_\pi V(s) = R(s) + \gamma \int_\mathcal{S} P(s'|s, \pi(s)) V(s') ds'$$

$$\bar{T} V(s) = \max_{\pi \in \Pi} \bar{T}_\pi V(s).$$

We additionally denote the Bellman operator for a policy which always selects action $a$ as

$$\bar{T}_a V(s) = R(s) + \int_\mathcal{S} P(s'|s, a) V(s') ds'.$$

If $\mathcal{S}$ is finite, the above Bellman operators can be expressed for all states using matrix notation, in which $V$ and $R$ are vectors of values and rewards for each state, respectively, and $P$ is a matrix containing probabilities of transitioning between each pair of states:

$$\bar{T}_\pi V = R_\pi + \gamma P_\pi V$$

$$\bar{T} V = \max_{\pi \in \Pi} \bar{T}_\pi V$$

$$\bar{T}_a V = R + \gamma P_a V$$

The optimal value function $V^*$ satisfies $\bar{T} V^* = V^*$.

Because $\bar{T}$ computes an exact expectation over next states, we refer to $\bar{T}$ as the Bellman operator *with expectation*. This requires *sampling with expectation*, defined as $\bar{\Sigma} \subseteq \{(s, a, \mathbb{E}[R(s)]), p(\cdot|s, a) | s \in \mathcal{S}, a \in \mathcal{A}\}$. This kind of sampling assumes that the complete density over next states is provided with every sample, a rather unrealistic assumption. It is therefore usually convenient to use *one-step simple samples* $\Sigma \subseteq \{(s, a, r, s')|s, s' \in \mathcal{S}, a \in \mathcal{A}\}$, where $s'$ is a single draw from $p(s'|s, a)$. An individual $(s, a, r, s')$ sample from sets $\bar{\Sigma}$ and $\Sigma$ will be denoted $\bar{\sigma}$ and $\sigma$, respectively. An individual element of a sample $\sigma$ will be denoted with superscripts; that is, the $s$ component of a $\sigma = (s, a, r, s')$ sample will be denoted $\sigma^s$. Simple samples have an associated *sampled* Bellman operator $T_\sigma$, where $T_\sigma V(\sigma^s) = \sigma^r + \gamma V(\sigma^{s'})$.

We will also define the optimal Q-function $Q^*(s, a)$ as the value of taking action $a$ at state $s$, and following the optimal policy thereafter. More precisely, $Q^*(s, a) = R(s, a) + \gamma \int_\mathcal{S} P(s'|s, a) V^*(s') ds'$.

We focus on *linear value function approximation* for discounted infinite-horizon problems, in which the value function is represented as a linear combination of *nonlinear basis functions (vectors)* which are assumed to be measurable on $S$. For each state $s$, we define a vector $\Phi(s)$ of features. The rows of the basis matrix $\Phi$ correspond to $\Phi(s)$, and the approximation space is generated by the columns of the matrix. That is, the basis matrix $\Phi$, and the value function $V$ are represented as:

$$\Phi = \begin{pmatrix} - & \Phi(s_1) & - \\ & \vdots & \end{pmatrix} \qquad V = \Phi w.$$

This form of linear representation allows for the calculation of an approximate value function in a lower-dimensional space, which provides significant computational benefits over using a larger or complete basis; if the number of features is small, this framework can also guard against overfitting any noise in the samples though small features sets alone cannot protect ALP methods from making errors due to noise.

## 3 $L_1$-Regularized Approximate Linear Programming

This section includes our discussion of $L_1$-Regularized Approximate Linear Programming. Subsection 3.1 will review the salient points of the method, while Subsection 3.2 will discuss the weaknesses of the method and our motivation for improvement.

### 3.1 RALP Background

RALP was introduced by Petrik et al. [12] to improve the robustness and approximation quality of Approximate Linear Programming (ALP) [13, 1]. For a given set of samples with expectation $\bar{\Sigma}$, feature set $\Phi$, and regularization pa-

rameter $\psi$, RALP calculates a weighting vector $w$ by solving the following linear program:

$$\begin{aligned}
\min_w \quad & \rho^T \Phi w \\
\text{s.t.} \quad & \bar{T}_{\sigma^a} \Phi(\sigma^s) w \leq \Phi(\sigma^s) w \quad \forall \sigma \in \Sigma \\
& \|w\|_{1,e} \leq \psi,
\end{aligned} \quad (1)$$

where $\rho$ is a distribution over initial states, and $\|w\|_{1,e} = \sum_i |e(i) w(i)|$. It is generally assumed that $\rho$ is a constant vector and $e = \mathbf{1}_{-1}$, which is a vector of all ones but for the position corresponding to the constant feature, where $e(i) = 0$.

Adding $L_1$ regularization to the constraints of the ALP achieved several desirable effects. Regularization ensured bounded weights in general and smoothed the value function. This prevented missing constraints due to unsampled states from leading to unbounded or unreasonably low value functions. By using $L_1$ regularization in particular, the well-known sparsity effect could be used to achieve automated feature selection from an extremely rich $\Phi$. Finally, the sparsity in the solution could lead to fast solution times for the LP even when the number of features is large because the number of active constraints in the LP would still be small.

As a demonstration of the sparsity effect, RALP was run on the upright pendulum domain (explained in detail in Section 6) using a set of features for each action consisting of three Gaussian kernels centered around each sampled state, one polynomial kernel based around each sampled state, and a series of monomials calculated on the sampled states. On one randomly-selected run of 50 sample trajectories (a trajectory consists of states reached as a result of randomly selected actions until the pendulum falls over), to represent the Q-values of the first action RALP selected zero of the narrowest Gaussian kernels, nine of the intermediate-width Gaussian kernels, five of the widest Gaussian kernels, five of the polynomial kernels, and zero of the monomials. For the second action, the Q value was represented with zero narrow, seven intermediate, and four wide Gaussian kernels, six polynomial kernels, and one monomial. For the final action, these numbers were one, four, four, seven, and zero. Selecting this same mixture of features by hand would be extremely unlikely.

Petrik et al. [12] demonstrated a guarantee that the error on the RALP approximation will be small, as will be further discussed in the next subsection. Additionally, in experiments, policies based on RALP approximations outperformed policies generated by Least-Squares Policy Iteration [9], even with very small amounts of data.

RALP is not the only approach to leverage $L_1$ regularization in value function approximation; others include LARS-TD [7], and LC-MPI [6], which both attempt to calculate the fixed point of the $L_1$-penalized LSTD problem. Unlike RALP, this solution is not guaranteed to exist when samples are drawn off-policy. These methods have the advantage of handling noise more gracefully than LP based approaches, but require on-policy sampling for reliable performance. On-policy sampling is often a challenging requirement for real-life applications.

### 3.2 RALP Weaknesses

The RALP constraints in LP 1 are clearly difficult to acquire – samples with expectation require a model that provides a distribution over next states, or access to a simulator that can be easily reset to the same states many times to compute an empirical next state distribution for each constraint.

Petrik et al. [12] previously presented approximation bounds for three constructions of RALP, depending upon the sampling regime. In the *Full ALP*, it is assumed every possible state-action pair is sampled with expectation. The *Sampled ALP* contains constraints for only a subset of state-action pairs sampled with expectation. The third form, the *Estimated ALP*, also has constraints defined on sampled state-action pairs, but of the form

$$\Phi(\sigma^s) w \geq T_\sigma \Phi(\sigma^s) w \quad \forall \sigma \in \Sigma.$$

Note that the Estimated ALP uses one-step simple samples and the sampled Bellman operator.

These three forms are presented in order of decreasing accuracy, but increasing applicability. The Full ALP requires a manageably small and discrete state space, as well as a completely accurate model of the reward and transition dynamics of the system. The Sampled ALP still requires the model, but allows for constraint sampling; this introduces constraint violation error (denoted $\epsilon_p$), as the approximation may violate unsampled constraints. Finally, the Estimated ALP requires neither the model nor experience at every state; because replacing an accurate model with a sample is likely to be imprecise in a noisy domain, this introduces model error (denoted $\epsilon_s$). However, this is the most realistic sampling scenario for learning problems that involve physical systems.

For reference, and to understand the effect of model error, we reproduce the RALP error bounds for the Sampled and Estimated ALP here. We denote the solution of the Sampled ALP $\bar{V}$, and the solution of the Estimated ALP $V$. $\epsilon_c$ is error from states that do not appear in the objective function because they were not samples. Since $\rho$ is often chosen fairly arbitrarily, we can assume the unsampled states have 0 weight and that $\epsilon_c = 0$.

$$\|\bar{V} - V^*\|_{1,\rho} \leq \epsilon + 2\epsilon_c(\psi) + 2\frac{\epsilon_p(\psi)}{1-\gamma} \quad (2)$$

$$\|V - V^*\|_{1,\rho} \leq \epsilon + 2\epsilon_c(\psi) + \frac{3\epsilon_s(\psi) + 2\epsilon_p(\psi)}{1-\gamma} \quad (3)$$

To understand the source of model error $\epsilon_s$, consider a case where a state-action pair is sampled twice. Further suppose that in one of these samples, the agent transitions to a high-value state; in the other, it transitions to a lower-valued state. The resulting value function is constrained by the high-valued sample, while the constraint related to the low-valued experience remains loose; that sample might as well have not occurred. The approximate value at that state will be inaccurately high. This scenario is depicted in Figure 1.

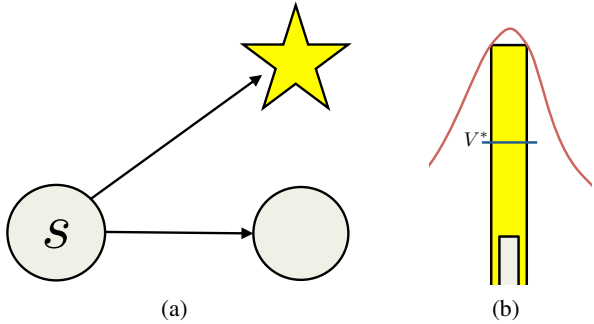

Figure 1: Subfigure 1(a) illustrates two samples beginning at state $s$; one transitions to the low-valued circle state, while the other transitions to the high-valued star state. Subfigure 1(b) illustrates the resulting constraints and value function approximation. $V^*$ is the actual value of this state, but the red line representing the approximate function is constrained by the max of the two constraints. Under realistic sampling constraints, it is more likely this scenario would occur with two samples drawn from very nearly the same state, rather than both originating from the same sample $s$.

The inability of LP approaches to value function approximation to handle noisy samples is well known. One proposed solution is Smoothed Approximate Linear Programming (SALP), in which the LP is given a budget with which to violate constraints [2]. Note, however, that there is nothing connecting the constraint violation budget to the problem dynamics unless additional information is provided. In Figure 1, for example, SALP could fortuitously choose the right amount of constraint violation to mix between the next state values for state $s$ in the proportions dictated by the model. However it is also possible for SALP to yield exactly the same value for this state as ordinary ALP, having spent its constraint violation budget on some other state which might not even be noisy.

## 4 Locally Smoothed $L_1$-Regularized Approximate Linear Programming

To address the weaknesses introduced in Subsection 3.2, we introduce Locally Smoothed $L_1$-Regularized Approximate Linear Programming (LS-RALP). The concept behind LS-RALP is simple; by averaging between nearby, similar samples, we can smooth out the effects of noise and produce a more accurate approximation. We define "nearby" to mean two things: one, the states should be close to each other in the state space, and two, the actions taken should be identical. The goal is to achieve the performance of the Sampled ALP with the more realistic sampling assumptions of the Estimated ALP.

We now introduce the smoothing function $k_b(\sigma, \sigma_i)$. A smoothing function spreads the contribution of each sampled data point $\sigma_i$ over the local neighborhood of data point $\sigma$; the size of the neighborhood is defined by the bandwidth parameter $b$. If $\sigma_i^a \neq \sigma^a$, the function returns zero. We also stipulate that $\sum_{\sigma_i \in \Sigma} k_b(\sigma, \sigma_i) = 1$ for all $\sigma$, and that $k_b(\sigma, \sigma_i) \geq 0$ for all $\sigma, \sigma_i$. Smoothing functions of this sort are commonly used to perform nonparametric regression; as sampled data are added, the bandwidth parameter is tuned to shrink the neighborhood, to allow for a balance between the variance from having too few samples in the neighborhood and the bias from including increasingly irrelevant samples which are farther away.

**Observation 4.1** *Consider the regression problem of estimating the function $f$ of the response variable $y = f(x) + N(x)$, given $n$ observations $(x_i, y_i)(i = 1, \cdots, n)$, where $N(x)$ is a noise function. Assume the samples $x_i$ are drawn from a uniform distribution over a compact set $G$, that $f(x)$ is continuously differentiable, and that $N(x)$ satisfies certain regularity conditions. There exists a smoothing function $k_b(x, x_i)$ and associated bandwidth function $b(n)$ such that $\forall x \in G$, as $n \to \infty, \sum_i k_{b(n)}(x, x_i)y_i \to f(x)$, almost surely.*

Smoothing functions that suffice to provide the consistency guarantees in the above observation include kernel estimators [3] and $k$-nearest-neighbor averagers [4]. Similar results exist for cases without uniform sampling [5, 3].

The regularity conditions on the noise function depend upon the type of estimator and allow for a variety of noise models [3, 4]. In some of these cases, other assumptions in Observation 4.1 can be weakened or discounted altogether.

Literature on nonparametric regression also contains extensive theory on optimal shrinkage rates for the bandwidth parameter; in practice, however, for a given number of samples, this is commonly done by cross-validation.

We now modify our LP to include our smoothing function:

$$\begin{aligned}
\min_w \quad & \rho^T \Phi w \\
\text{s.t.} \quad & \sum_{\sigma_i \in \Sigma} k_b(\sigma, \sigma_i) T_{\sigma^a} \Phi(\sigma_i^s) w \leq \Phi(\sigma_i^s) w \quad \forall \sigma \in \Sigma \\
& \|w\|_{1,e} \leq \psi.
\end{aligned} \quad (4)$$

In this formulation, we are using our smoothing function to estimate the constraint with expectation $\bar{T}\Phi(\sigma^s)w$ by smoothing across our easily obtained $T_\sigma \Phi(\sigma^s)w$.

Note that unsmoothed RALP is a special case of LS-RALP, where the bandwidth of the smoothing function is shrunk until the function is a delta function. Therefore, LS-RALP, with correct bandwidth choice, can always do at least as well as RALP.

We note that smoothing functions have been applied to reinforcement learning before, in particular by Ormoneit and Sen [11], whose work was later extended to policy iteration by Ma and Powell [10]. In this formulation, a kernel estimator was used to approximate the value function; this differs significantly from our approach of using a smoother to calculate weights more accurately for a parametric approximation. Additionally, because their approximation simply averages across states and does not use features, it will tend to have a poor approximation for novel states unless the space is very densely sampled. More recently, and independently from our work, Kroemer and Peters [8] proposed an approach to policy evaluation that first used kernel density estimation to approximate the dynamics and then solved for the value function of resulting approximate system using dynamic programming.

## 5 Convergence Proof

The goal of this section is to demonstrate that the application of a smoothing function to the constraints of the Estimated ALP will mitigate the effects of noise. In particular, we show that as the number of samples approaches infinity, the LS-RALP solution using one-step samples approaches the RALP solution using samples with expectation.

We begin by providing an explicit definition of $\epsilon_s$ that is consistent with the usage in Petrik et al. [12]. For a set of samples $\Sigma$,

$$\epsilon_s = \max_{\sigma \in \Sigma} |\bar{T}\Phi(\sigma^s)w - T_\sigma \Phi(\sigma^s)w|. \quad (5)$$

We now introduce and prove our theorem. We denote the number of samples $|\Sigma|$ as $n$.

**Theorem 5.1** *If the reward function, transition function, and features are continuously differentiable, samples are drawn uniformly from a compact state space, and the noise models of the reward and transition functions satisfy the conditions of Observation 4.1 that would make kernel regression converge, then there exists a bandwidth function $b(n)$ such that as $n \to \infty$, the LS-RALP solution using one-step samples approaches the Sampled RALP solution for all states in the state space.*

We begin by restating the definition of $\epsilon_s$:

$$\epsilon_s = \max_{\sigma \in \Sigma} |\bar{T}\Phi(\sigma^s)w - T_\sigma \Phi(\sigma^s)w|.$$

We then expand upon the portion within the max operator for some arbitrary $\sigma$:

$$\bar{T}\Phi(\sigma^s)w - T_\sigma\Phi(\sigma^s)w$$
$$= \left[R(\sigma^s) + \gamma \int_\mathcal{S} P(ds'|\sigma^s, \sigma^a)\Phi(s')w\right] - \left[\sigma^r + \gamma\Phi(\sigma^{s'})w\right]$$

We now introduce the smoothing function so as to match the constraints of LS-RALP; that is, we replace $T_\sigma\Phi(\sigma^s)w$ with $\sum_{\sigma_i \in \Sigma} k_b(\sigma, \sigma_i) T_\sigma \Phi(\sigma_i^s) w$.

$\epsilon_s$ remains the maximum difference between the ideal constraint $\bar{T}\Phi(\sigma^s)w$ and our existing constraints, which are now $\sum_{\sigma_i \in \Sigma} k_b(\sigma, \sigma_i) T_\sigma \Phi(\sigma_i^s) w$.

$$\bar{T}\Phi(\sigma^s)w - \sum_{\sigma_i \in \Sigma} k_b(\sigma, \sigma_i) T_\sigma \Phi(\sigma_i^s) w$$
$$= \left[R(\sigma^s) + \gamma \int_\mathcal{S} P(ds'|\sigma^s, \sigma^a)\Phi(s')w\right]$$
$$- \sum_{\sigma_i \in \Sigma} k_b(\sigma, \sigma_i) \left[\sigma_i^r + \gamma \Phi(\sigma_i^{s'})w\right]$$
$$= R(\sigma^s) - \sum_{\sigma_i \in \Sigma} k_b(\sigma, \sigma_i) \sigma_i^r$$
$$+ \gamma \int_\mathcal{S} P(ds'|\sigma^s, \sigma^a)\Phi(s')w - \gamma \sum_{\sigma_i \in \Sigma} k_b(\sigma, \sigma_i) \Phi(\sigma_i^{s'})w$$

The smoothing function is performing regression on all observed data, rather than just the single, possibly-noisy sample. The first two terms of the above equation are the difference between the expected reward and the predicted reward, while the second two are the difference between the expected next value and the predicted next value. The more accurate our regression on these two terms, the smaller $\epsilon_s$ will be.

It follows from Observation 4.1 that as $n$ increases, if $R(s), P(s'|s, a)$, and $\Phi(s)$ are all continuously differentiable with respect to the state space, and if samples are drawn uniformly from a compact set, and the noise model of the reward and transition functions is one of several allowed noise models, there exists a bandwidth shrinkage function $b(n)$ such that $\sum_{\sigma_i \in \Sigma} k_b(\sigma, \sigma_i) T_\sigma \Phi(\sigma_i^s) w$ will approach $\bar{T}\Phi(\sigma^s)w$ almost surely, and $\epsilon_s$ will approach 0 almost surely.

When $\epsilon_s = 0$, the smoothed constraints from one-step samples of LS-RALP are equivalent to the constraints with expectation of the Sampled RALP. ∎

The assumptions on $R(s), P(s'|s, a)$, and $\Phi(s)$ required for our proof may not always be realistic, and are stronger than the bounded change assumptions made by RALP. However, we will demonstrate in the next section that meeting these assumptions is not necessary for the method to be effective.

This theorem shows that sample noise, the weak point of all ALP algorithms, can be addressed by smoothing in a principled manner. Even though these results address primarily the limiting case of $n \to \infty$, our experiments demonstrate that in practice we can obtain vastly improved value functions even with very few samples.

## 6 Experiments

In this section, we apply LS-RALP to noisy versions of common reinforcement learning benchmark problems to demonstrate the advantage of smoothing. We will use two domains, the inverted pendulum [15] and the mountain-car [14]. While we have proved that LS-RALP will improve upon the value function approximation of RALP as the number of samples grows large, our experiments demonstrate that there is significant improvement even with a relatively small number of samples.

In the mountain-car domain, we approximate the value function, and then apply the resulting greedy policy. We generate the policy via the use of a model; at each state, we evaluate each candidate action, and calculate the approximate value at the resulting state. The action which resulted in the highest value is then chosen in the trajectory.

In the pendulum problem, we extend the method to the approximation of Q-functions, and again apply the resulting greedy policy. When using Q-functions, the model is unnecessary; the action with the highest Q-value is chosen.

### 6.1 Mountain Car

In the mountain car problem, an underpowered car must climb a hill by gaining momentum. The state space is two-dimensional (position and velocity), the action space is three-dimensional (push, pull, or coast), and the discount factor is 0.99. Traditionally, a reward of $-1$ is given for any state that is not in the goal region, and a reward of 0 is given for any state that is. For this problem we applied Gaussian noise with standard deviation .5 to the reward to demonstrate the ability of LS-RALP to handle noisy rewards. This is a very large amount of noise that greatly increases the difficulty of the problem.

Features are the distributed radial basis functions suggested by Kolter and Ng [7]; however, because LSPI with LARS-TD requires a distinct set of basis functions for each action, while RALP methods do not, our dictionary is one-third of the size. For sake of a baseline comparison, we note that in the deterministic version of this problem, we repeated experiments by Kolter and Ng [7] and Johns et al. [6] on LARS-TD and LC-MPI, respectively and RALP outperformed both.

Sample trajectories started at a random state, and were allowed to run with a random policy for 20 steps, or until the goal state was reached. After the value function was calculated, it was tested by placing the car in a random state; the policy then had 500 steps for the car to reach the goal state.

Unlike the paper which introduced RALP [12], only one action at each state was sampled.

Figure 2 shows the success rate over 50 trials for both RALP and LS-RALP. The regularization parameter for the RALP methods was 1.4, and the smoothing function for LS-RALP was a multivariate Gaussian kernel with a standard deviation spanning roughly 1/85 of the state space in both dimensions. These parameters were chosen using cross validation.

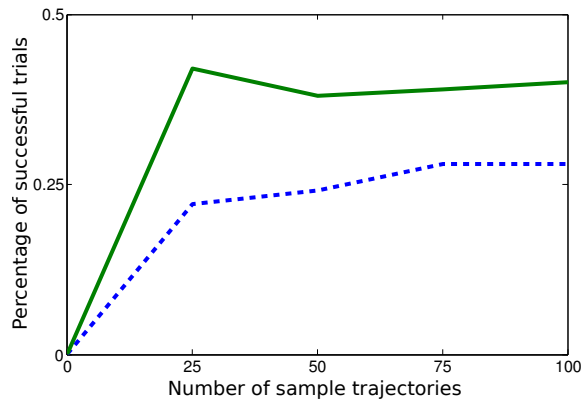

Figure 2: Percentage of successful trials vs. number of sample trajectories for the mountain-car domain with value functions approximated by RALP (blue, dotted) and LS-RALP (green, solid).

### 6.2 Inverted Pendulum

The focus of this paper is value function approximation, but our theory extends trivially to Q-function approximation. We demonstrate this by using RALP and LS-RALP to approximate Q-functions; this completely removes the need for a model in policy generation.

In the inverted pendulum problem the agent tries to balance a stick vertically on a platform by applying force to the platform. The state space is the angle and angular velocity of the stick, and the action space contains three actions: a push on the table of 50N, a pull on the table of 50N, and no action. The discount factor is 0.9. The goal is to keep the pendulum upright for 3000 steps. The noise in the pendulum problem is applied as Gaussian noise of standard deviation 10 on the force applied to the table. This is significantly greater than the noise traditionally applied in this problem, resulting in much worse performance even by the best controller we could produce.

We compare RALP and LS-RALP, using the same set of features; Gaussian kernels of three different widths, a polynomial kernel of degree 3, $s[1], s[2], s[1] * s[2], s[1]^2, s[2]^2$, and a constant feature, where $s[1]$ is the angle, and $s[2]$ is the angular velocity. For a number of samples $n$, this was therefore a total of $4n + 6$ features for each action. We also compare against LSPI, using the radial basis functions used by Lagoudakis and Parr [9], and LARS-TD wrapped with policy iteration, using the same features used by RALP and LS-RALP.

Sampling was done by running 50 trajectories under a random policy, until the pendulum fell over (around six steps). For the LP methods, the regularization parameter was set to 4000, and for LS-RALP, the smoothing function was a multivariate Gaussian kernel. These parameters were chosen by cross validation. 50 trials were run for each method, and the number of steps until failure averaged; if a trial reached 3000 steps, it was terminated at that point. As with the mountain car experiments, only one action for each sampled state was included. Results are presented in Figure 3.

Notice that while LSPI was competitive with the Q-functions approximated with RALP, the addition of the smoother resulted in superior performance even with very little data.

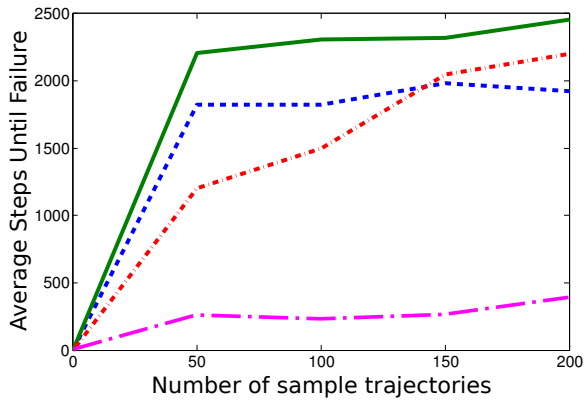

Figure 3: Average number of steps until failure vs. number of sample trajectories for the pendulum domain with Q-functions approximated with LARS-TD with policy iteration (violet, hashed), LSPI (red, hashed), RALP (blue [dark gray], dotted), and LS-RALP (green, solid).

## 7  Conclusion

In this paper, we have demonstrated the difficulties RALP can have in using noisy, sampled next states, and we have shown that the introduction of a smoothing function can mitigate these effects. Furthermore, we proved that as the amount of data is increased, error from inaccurate samples approaches zero. However, our experiments show drastic improvement with only a small number of samples.

In fairness, we also point out some limitations of our approach: LS-RALP makes stronger smoothness assumptions about the model and features than does RALP (continuous differentiability vs. the Lipschitz continuity necessary for the RALP bounds). LS-RALP also requires an additional parameter that must be tuned - the kernel bandwidth. We note, however, that the RALP work provides some guidance on choosing the regularization parameter and that there is a rich body of work in non-parametric regression that may give practical guidance on the selection of the bandwidth parameter.

The nonparametric regression literature notes that the bandwidth of a smoother can be adjusted not just for the total number of data points, but also for the density of data available in the particular neighborhood of interest [5]. The addition of this type of adaptiveness would greatly increase accuracy in areas of unusual density.

There are further opportunities to make better use of smoothing. First, we note that while smoothing the entire constraint has proven to be extremely useful, it is inevitable that applying two separate smoothing parameters to the reward and transition components would be even more helpful. For example, in the mountain-car experiments of Subsection 6.1, only the reward was noisy; applying smoothing to the reward, and not to the transition dynamics, would improve the approximation. The introduction of an automated technique of setting bandwidth parameters on two separate smoothing functions would likely result in more accurate approximations with less data.

Additionally, while the experiments done for this paper were extremely quick, this was due to the small number of samples needed. For larger, more complex domains, the LP can get large and may take a long time to solve; while LP solvers are generally thought of as efficient, identification of constraints doomed to be loose or features which will not be selected could cause a drastic improvement in the speed of this method.

## Acknowledgments

This work was supported in part by NSF IIS-0713435 and NSF IIS-1147641, as well as the Naval Academy Research Council. Opinions, findings, conclusions or recommendations herein are those of the authors and not necessarily those of the sponsors.